\documentclass{article}

\usepackage{microtype}
\usepackage{graphicx}
\usepackage{subfigure}
\usepackage{booktabs} 
\usepackage{amsmath}
\usepackage{amssymb}
\usepackage{caption}
\usepackage{paralist}
\usepackage{scalefnt}
\usepackage{xspace}
\usepackage{natbib}
\usepackage{comment}

\usepackage[accepted]{icml2018}

\icmltitlerunning{Accuracy-based Curriculum Learning in deep RL}

\newcommand{\ddpg}{\textsc{ddpg}\xspace}
\newcommand{\uvfa}{\textsc{uvfa}\xspace}

\newcommand{\dqn}{\textsc{dqn}\xspace}

\newcommand{\her}{{\sc her}\xspace}

\newcommand{\abcl}{accuracy-based curriculum learning\xspace}

\definecolor{myred}{rgb}{0.8,0,0}
\definecolor{mygreen}{rgb}{0,0.6,0}
\definecolor{myblue}{rgb}{0,0,0.7}

\begin{document}


\twocolumn[
\icmltitle{Accuracy-based Curriculum Learning in Deep Reinforcement Learning}


\begin{icmlauthorlist}
\icmlauthor{Pierre Fournier}{su}
\icmlauthor{Mohamed Chetouani}{su}
\icmlauthor{Pierre-Yves Oudeyer}{in}
\icmlauthor{Olivier Sigaud}{su,in}
\end{icmlauthorlist}

\icmlaffiliation{su}{Sorbonne Universit\'e, ISIR, Paris, France}
\icmlaffiliation{in}{INRIA, Flowers Team, Bordeaux, France}

\icmlcorrespondingauthor{Pierre Fournier}{pierre.fournier@isir.upmc.fr}

\icmlkeywords{deep reinforcement learning, curriculum learning, continuous actions}

\vskip 0.3in
]

\newcommand{\fix}{\marginpar{FIX}}
\newcommand{\new}{\marginpar{NEW}}


\printAffiliationsAndNotice{}  

\begin{abstract}
In this paper, we investigate a new form of automated curriculum learning based on adaptive selection of accuracy requirements, called \abcl.
Using a reinforcement learning agent based on the Deep Deterministic Policy Gradient algorithm and addressing the Reacher environment, we first show that an agent trained with various accuracy requirements sampled randomly learns more efficiently than when asked to be very accurate at all times. Then we show that adaptive selection of accuracy requirements, based on a local measure of competence progress, automatically generates a curriculum where difficulty progressively increases, resulting in a better learning efficiency than sampling randomly.
\end{abstract}


\section{Introduction}

When an agent has to learn to achieve a set of tasks, the curriculum learning problem can be defined as the problem of finding the most efficient sequence of learning situations in these various tasks so as to maximize its learning speed, either over the whole set of tasks or with respect to one of these tasks.

A rule of thumb in curriculum learning is that one should address easy tasks first and switch to more difficult tasks once the easy ones are correctly mastered, because competences obtained on learning from the easy ones may facilitate learning on the more difficult ones, through {\em transfer learning}. 
However, as machine learning algorithms have complex biases, it is often difficult to design by hand an efficient learning curriculum. Also, various task sets, as well as various learners, may differ in terms of which are the best learning curriculum. An important scientific challenge is thus how to design algorithms that can incrementally and online generate a curriculum that is efficient for a specific set of tasks and a specific learner. 

An idea that has been explored in various strands of the literature \citep{schmidhuber1991curious,oudeyer2007intrinsic} has been to generate a learning curriculum by dynamically selecting learning situations which provide maximal learning progress at a given point in time. This idea has been used to automate the generation of learning curriculum for training robots \citep{baranes2013active}, deep neural networks \citep{graves2017automated}, as well as human learners in educational settings \citep{clement2013multi}. It entails several challenges, including how to efficiently estimate expected learning progress, and how to optimize exploration and exploitation to maximize learning progress with bandit-like algorithms. 

Another important challenge entailed by this approach is how to parameterize learning situations, i.e. to design the parameters over which learning progress is estimated and compared.  
In the case of an agent learning to reach various goals in the environment, the general notion of learning progress which might for instance denote progress in predicting some signal is replaced by a narrower notion of competence progress, 
which is more specific to learning to act. In the case of reaching, a natural parameterization consists in defining learning situations as particular regions of goal states (e.g. \citep{baranes2013active}),
or particular regions of starting states (e.g. \citep{florensa2017reverse}), or particular combinations of starting and end states. As some target or starting states might be easier to learn than others, thus producing higher competence progress in the beginning, a strategy based on competence progress will first focus on them and then move towards more complicated ones. If this parameterization is continuous, architectures like SAGG-RIAC \citep{baranes2013active} can be used to dynamically and incrementally learn these regions, and concurrently use them to select and order learning situations.

In this paper, we focus on another way to parameterize learning situations based on the notion of accuracy requirement of the tasks/goals. 
Many robotics tasks can be made more or less difficult by requiring different degrees of accuracy from the robot: learning to bring its end-effector within 10cm of a point may be easier than within 10mm.
A task which was easy with loose accuracy requirements can become difficult if the accuracy constraint becomes tighter. 
The impact of accuracy requirement on learning efficiency in the context of curriculum learning can be particularly powerful, since in reinforcement learning (RL) progress is made by finding rewarding experiences, and it is harder to find a source of reward when the accuracy requirement is stronger.  

To capture this idea, we consider the required accuracy $\epsilon$ as a parameter of learning situations, and we define accuracy-based curriculum learning as a specific form of curriculum learning which acts on the value of $\epsilon$ to improve learning efficiency. Then we study the impact of \abcl on the learning efficiency of a multi-goal deep RL agent trying to learn various reaching tasks.

Our contributions are the following. First, we show that being trained with various accuracy requirements is beneficial to learning efficiency even when the required accuracies are sampled in a random order. Then we show that using \abcl based on the competence progress as defined in \citep{baranes2013active} enables to automate the sampling of accuracy constraints in order of increasing difficulty, and that such ordering results in a better learning efficiency than random sampling.

The paper is organized as follows. In Section~\ref{sec:related}, we present some works that are related to our main concerns. 
In Section~\ref{sec:methods}, we quickly describe the Reacher environment used as experimental setup, the deep RL algorithm, Deep Deterministic Policy Gradient (\ddpg)  \citep{lillicrap2015continuous} and its Universal Value Function Approximation (\uvfa) extension, as well as the \abcl algorithm we are using. 
We describe our results in Section~\ref{sec:results}. Finally, we discuss these results, conclude and describe potential avenues for future work in Section~\ref{sec:conclusion}.

\section{Related work}
\label{sec:related}

Our work contributes to the domain of curriculum learning, leveraging contributions made in different fields.

One line of research in machine learning \citep{schmidhuber1991curious} has proposed that measures of improvement of prediction errors
could be used as intrinsic rewards to be maximized, generating a learning curriculum in an intrinsically motivated RL framework \citep{chentanez2005intrinsically}. It was later on extended to drive the selection of problems in the Powerplay framework \citep{schmidhuber2013powerplay}.

Another independent line of research, in the domain of developmental robotics, has focused on modeling children's curiosity-driven exploration
as a mechanism that enables them to learn world models through the adaptive selection of learning situations that maximize learning progress \citep{oudeyer2007intrinsic}. This line of research has developed algorithms enabling learning progress to be efficiently estimated in large continuous high-dimensional spaces \citep{baranes2009r}, enabling robots to self-organize a learning curriculum and learn incrementally and online complex skills such as locomotion \citep{baranes2013active}, tool use \citep{forestier2016curiosity} or soft object manipulation \citep{nguyen2014socially}. This line of research also explored a variety of ways in which learning progress could be used to select learning situations, including the automated selection of policy parameters \citep{oudeyer2007intrinsic}, goals \citep{baranes2013active}, models \citep{forestier2016curiosity}, and learning strategies \citep{nguyen2012active}. While some of these lines of research have also considered the possibility to progressively increase forms of perceptual accuracy based on learning progress (e.g. the McSAGG-RIAC architecture, \citep{oudeyer2013intrinsically}), they have not considered the possibility to dynamically switch between different accuracy requirements levels as we study in this article. 

These ideas were then introduced together in the context of deep neural network research in \cite{bengio2009curriculum}, which popularized the term "curriculum learning". They are now the focus of intensive research efforts both in developmental robotics \citep{forestier2016curiosity,pere2018unsupervised} and deep learning research \citep{graves2017automated,matiisen2017teacher,florensa2017reverse}, and are key elements of recent curiosity-driven deep reinforcement learning techniques enabling to solve RL problems with rare or deceptive rewards, e.g. \citep{bellemare2016unifying, machado2017laplacian}.
They are also the focus of some theoretical work \citep{weinshall2018curriculum}, for example to characterize the families of task sets for which the learning progress mechanism generates an optimal curriculum \citep{lopes2012strategic}.

In this context, our work follows closely the multi-goal architecture from the {\em Hindsight Experience Replay} (\her) algorithm \citep{andrychowicz2017hindsight}. The key innovation in \her consists in letting the agent learn again from previous experience by replaying real transitions of an episode with different goals reached during that episode. The algorithm is built on a combination of Universal Value Function Approximators (\uvfa) \citep{schaul2015universal}
and the \ddpg \citep{lillicrap2015continuous} deep RL algorithm. It is shown to provide the agent with a form of implicit curriculum learning based on target goal positions. The present work is also built on \uvfa{}s and \ddpg but unlike \her, it relies on a principled idea of competence progress to provide an explicit curriculum for the agent. To our knowledge, our work is the first to combine \uvfa{}s, \ddpg and an explicit measure of competence progress.

\section{Methods}
\label{sec:methods}

The idea of building a curriculum of tasks on the base of various accuracy requirements can be applied to many environments and learning algorithms, provided that the latter can incorporate such requirements. In this paper we apply it to solve the simple Reacher environment from OpenAI Gym \citep{brockman2016openai} with the \ddpg algorithm.

\subsection{Reacher}
\label{sec:reacher}

The reacher environment is a two degree of freedom arm that must reach a target in its plane. In this environment, the agent state contains the arm angles and angular velocities, its end-effector position $p_{finger}$ and the target position $p_{target}$. We use a sparse reward function that happens to naturally incorporate a notion of accuracy:
\begin{equation}
r =
    \begin{cases}
      0, & \text{if}\ |p_{finger}-p_{target}| \leq \epsilon \\
      -1, & \text{otherwise}
    \end{cases}
\end{equation}
with the parameter $\epsilon$ determining the accuracy required from the agent for being successful.

Importantly, a new target position $p_{target}$ is randomly sampled after each trial, providing a natural form of exploration.

\subsection{\ddpg}
\label{sec:ddpg}

The \ddpg algorithm \citep{lillicrap2015continuous} is a deep RL algorithm based on the Deterministic Policy Gradient \citep{silver2014deterministic}. It borrows the use of a replay buffer and a target network from \dqn \citep{mnih2015human}. 

The implementation of \ddpg used in this work differs from the original one by several aspects. First, we use the gradient inversion technique from \cite{hausknecht2015deep} to better deal with the bounded action space and to avoid saturating the final $tanh$ layer of the actor. Second, we use target clipping from \cite{andrychowicz2017hindsight} to mitigate target critic overestimation.

All the other parameters of \ddpg used in this paper are taken from baselines \citep{baselines} used in systematic studies \citep{islam2017reproducibility,henderson2017deep}: we use a minibatch of size 64, a replay buffer of size $1e6$, the critic and actor learning rates are respectively $0.001$ and $0.0001$, $\tau = 0.001$ and $\gamma = 0.99$. Both critic and actor networks have two hidden layers of 400 and 300 neurons with ReLU activations and are densely connected. They are trained with the Adam optimizer.

In the experiments below, we do not use any exploration noise, as we have determined that such exploration noise is detrimental to performance in the Reacher environment, due to the intrinsic exploration provided by the environment itself.

\subsection{Universal Value Function Approximators}
\label{sec:uvfa}

In the Reacher environment, the state contains both the robot position $p_{finger}$ and the target position $p_{target}$ and the target position is sampled randomly, making it natural to consider \ddpg as implementing a form of Universal Value Function Approximator (\uvfa) \citep{schaul2015universal}.

However, we use a slightly more complex representation than just this state.
Indeed, in the standard setting for RL with sparse rewards, the $\epsilon$ parameter is set at the beginning of training to a user-defined value and left untouched afterwards, whereas here $\epsilon$ changes from one training episode to the next. But changing $\epsilon$ during the reward/termination calculation without adding its value to the state would remove the MDP structure of the problem: a given state could be either terminal or not depending on $\epsilon$, leading to incoherent updates for \ddpg.
Thus we extend the \uvfa framework to contain both states and accuracies using a $V(s,\epsilon,\theta)$ representation. The MDP structure of the environment is not changed by this addition: $\epsilon$ simply keeps constant whatever the transition, and is only used to compute the termination condition. 

\subsection{Epsilon-based curriculum}
\label{sec:accuracy_curriculum}

Two training strategies are studied in this work. The first, called \textsc{random-$\epsilon$}, consists in sampling a training accuracy $\epsilon$ uniformly from $E=\{0.02, 0.03, 0.04, 0.05\}$ at the beginning of each episode, and adding it to the input state of the \uvfa for all the duration of the episode.

The second, called \textsc{active-$\epsilon$}, consists in measuring the agent competence progress for each $\epsilon \in E$ all along training, and sampling an accuracy at the beginning of each episode proportionally to its current competence progress measurement. Specifically, if $cp_i$ is the competence progress for $\epsilon_i$, then we define the probability of sampling $\epsilon_i$ as 
\begin{equation}
P(\epsilon_i) = \frac{cp_i^{\beta}}{\sum_{k} cp_k^{\beta}}.
\end{equation}

The exponent $\beta$ determines how much prioritization is used, with $\beta = 0$ corresponding to the uniform sampling case (\textsc{random-$\epsilon$}), and $\beta = \infty$ to only sampling the accuracy value $\epsilon$ resulting in the maximum competence progress. In the experiments below, we determined by sampling $\beta$ in the range $[0,8]$ that using a value of $\beta = 4$ was providing the best results.

Both strategies are compared to a baseline that corresponds to the \ddpg algorithm being trained to reach target positions with rewards obtained from the hard accuracy $\epsilon = 0.02$. Thus the baseline is always trained with accuracy constraints that correspond to that used for evaluation. By contrast, both \textsc{active-$\epsilon$} and \textsc{random-$\epsilon$} are designed to sometimes train on easier requirements, and thus train less with the evaluation accuracy constraint.

\subsection{Competence progress measurement}
\label{sec:learning progress}

The way we measure learning progress is directly taken from the SAGG-RIAC algorithm \citep{baranes2013active}.

For each possible value of $\epsilon \in E$, the agent is tested every 1000 steps on its ability to reach 10 randomly sampled target positions with this accuracy, providing a competence score between 0 and 1. From the ordered list of scores obtained this way during training,  we extract a measure of competence progress following \cite{baranes2013active}: if $c_k^i$ is the score measured for $\epsilon_i$ at measurement time $t_k$, then the competence progress $cp_i$ at time $T$ is the discrete absolute derivative of competence scores for $\epsilon_i$ over a sliding window of the $2N$ more recent evaluations:
\begin{equation}
cp_i = \frac{\left| \left(\displaystyle\sum_{j=T-N}^{T} c_j^i \right) - \left(\displaystyle\sum_{j=T-2N}^{T-N} c_j^i \right) \right|}{2N}.
\end{equation}

The use of an absolute value guarantees that if competence starts dropping for a given $\epsilon$, then its associated progress will be negative with high absolute value and thus will be sampled in priority over other slowly progressing accuracies. This avoids catastrophic forgetting. For all experiments, $N=3$ is used to compute progress on sliding windows containing 3000 steps.

\section{Results}
\label{sec:results}

In this Section we validate the hypotheses that: (1) sampling $\epsilon$ randomly from $E$ is more efficient for learning high accuracy behaviors than always using the strongest accuracy requirement ($\epsilon = 0.02$), and this despite seeing less experience with such a high accuracy; (2) sampling $\epsilon$  using competence progress on each accuracy level results in ordering accuracy requirements from the easiest to the most difficult; (3) the resulting curriculum improves learning efficiency over random selection.
%
%

\begin{figure}[t!]
	\includegraphics[width=\linewidth]{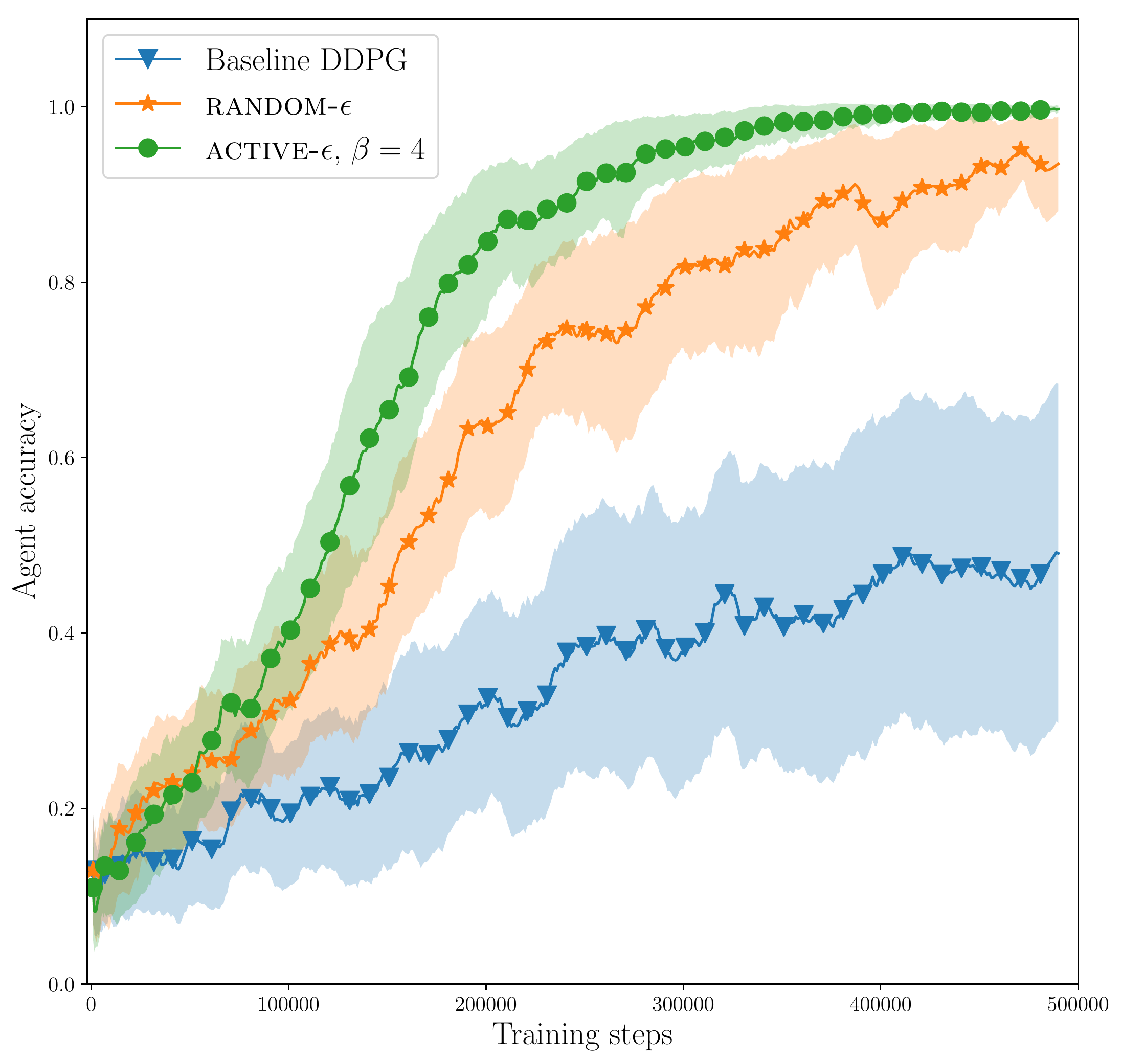}
	\caption{Average observed accuracy of the agent on 10 randomly sampled target positions across the environment with a required accuracy of 0.02 versus number of steps taken, averaged over 10 independent runs. The \textsc{random-$\epsilon$} and \textsc{active-$\epsilon$} (with $\beta = 4$) strategies are compared to the baseline algorithm alone. Colored areas cover one standard deviation around the mean of the 10 runs.}
	\label{fig:accuracy}
\end{figure}

Figure~\ref{fig:accuracy} shows the benefits from training with multiple accuracies as well as that of prioritizing values of $\epsilon$ depending on the level of competence progress of the agent. Every 1000 steps, the evaluation consists in running the agent for ten 50-step episodes on randomly sampled target positions, and recording the proportion of successful episodes for $\epsilon = 0.02$.
%
%

The \textsc{random-$\epsilon$} strategy already provides a clear gain over the baseline: one can observe a significant increase in accuracy as well as a reduction of the variance across different runs. The \textsc{active-$\epsilon$} strategy is shown with $\beta = 4$. The resulting curriculum provides even better learning performance than \textsc{random-$\epsilon$}: progress is faster at the beginning of the learning curve, the agent is more accurate in the end and shows less variability.

\begin{figure}[t!]
	\includegraphics[width=\linewidth]{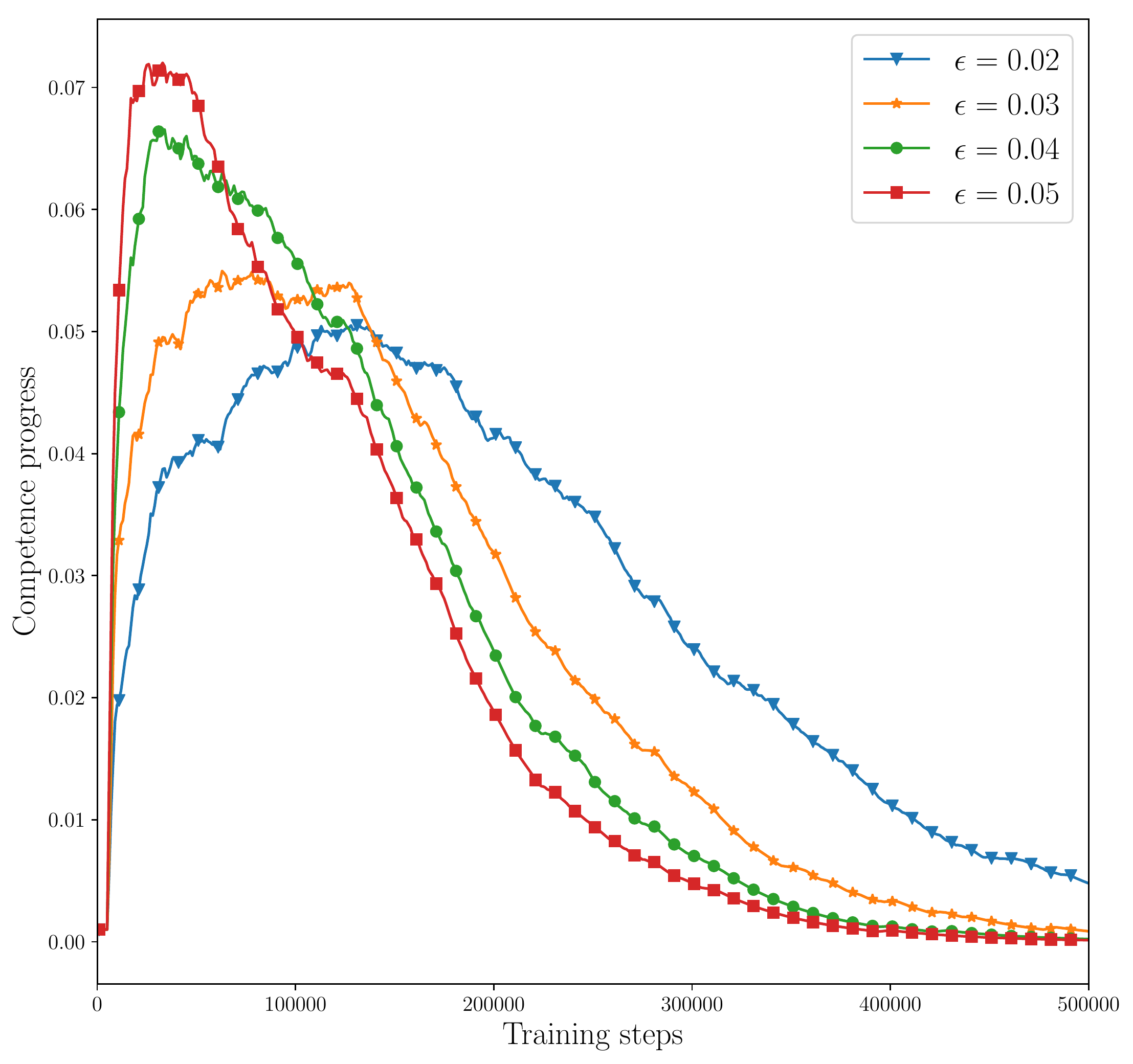}
	\caption{Evolution of competence progress for each $\epsilon \in E=\{0.02, 0.03, 0.04, 0.05\}$ with training steps, averaged over 10 runs, for the \textsc{active-$\epsilon$} strategy with $\beta=4$.}
	\label{fig:CP}
\end{figure}

\begin{figure}[t!]
	\includegraphics[width=\linewidth]{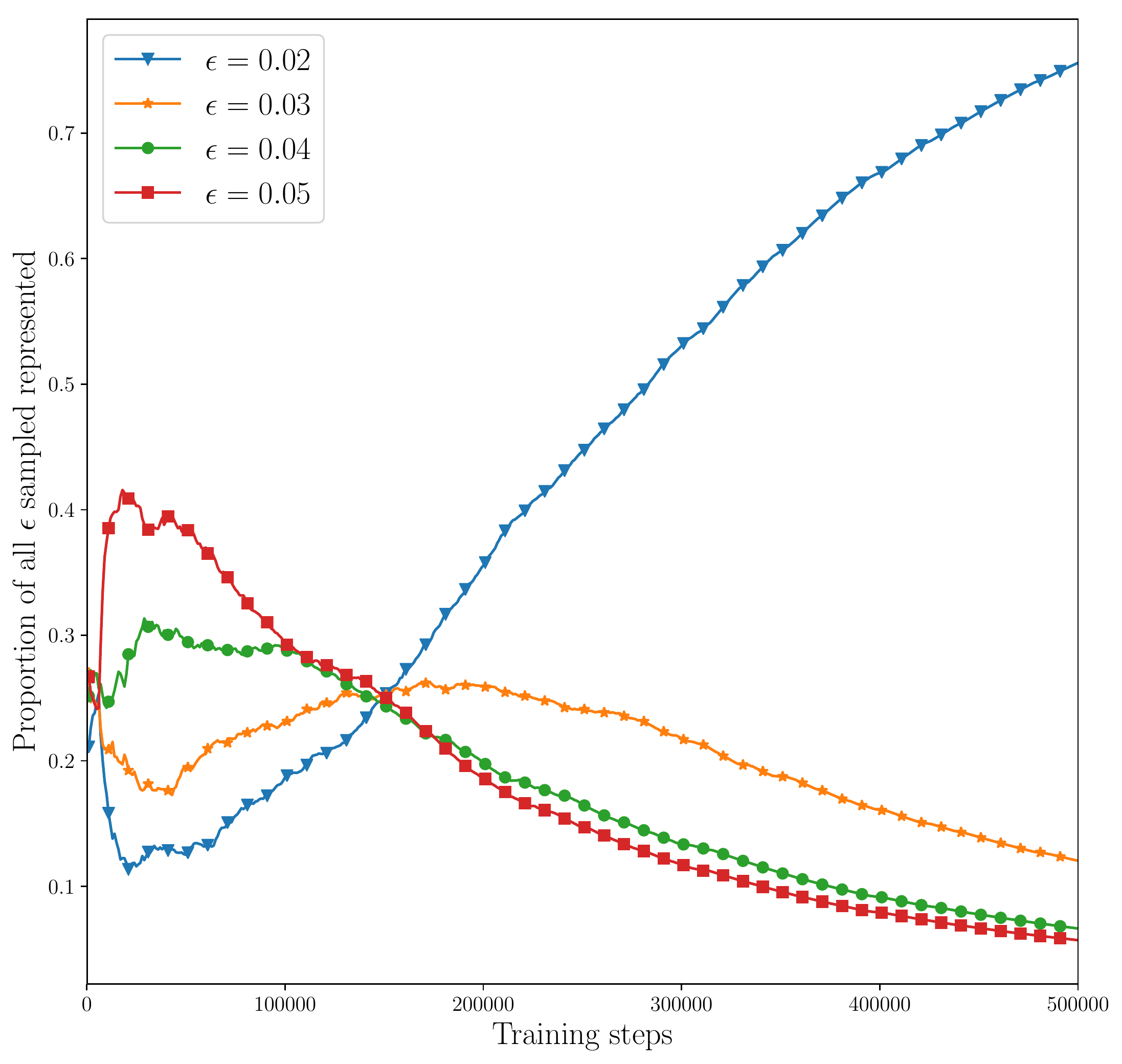}
	\caption{Proportions of all $\epsilon$ sampled during training represented by each specific value in $E$, versus number of steps taken, averaged over 10 runs, for the \textsc{active-$\epsilon$} strategy with $\beta=4$.}
	\label{fig:freq}
\end{figure}

Figures~\ref{fig:CP} and \ref{fig:freq} focus on the curriculum obtained with $\beta=4$ and parallels the evolution of competence progresses for each value of $\epsilon$ with their sampling frequencies. We observe on Fig.~\ref{fig:CP} that lower precisions lead to quicker progress at the beginning of learning compared to the most demanding task with $\epsilon = 0.02$. After about 150K steps, the opposite trend is observed: the agent competence on low precisions starts to reach a plateau as it masters the reaching task with these accuracies, leading to a decrease in progress; instead high accuracy remains challenging and the associated competence progress stays higher until later during training. 

In parallel, Fig.~\ref{fig:freq} shows the proportions of all $\epsilon$ sampled represented by each specific value of $\epsilon$, and reflects how their sampling frequency changes with training. During the first 150K steps priority is given to low accuracy objectives with high competence progresses, and the situation is reverted afterwards, with the agent almost only sampling the strongest accuracy requirement in the end, for which it is still making progress. 

\section{Discussion and conclusion}
\label{sec:conclusion}

We have shown that simply training on multiple accuracies and thus seeing \emph{less} experience with $\epsilon = 0.02$ in favor of larger $\epsilon$ values provides a substantial gain over the baseline. This suggests that the agent is able to take advantage of rewarding experience acquired with large $\epsilon$ values and then properly generalize the policy learned from these values to smaller ones. This idea is independent of the Reacher environment and could be leveraged to many simulated environments such as the Fetch environment from OpenAI \citep{brockman2016openai} or even to real robots.
However, this result may depend from a specific feature of accuracy-based difficulty, which is that a trajectory fulfilling some accuracy requirement also fulfills any easier accuracy requirement. Difficulties in terms of final points do not share a similar structure, for instance.
%
%

Also we demonstrated that competence progress is an effective metric to build a curriculum in the context of deep reinforcement learning, and could be used with groups of tasks that differ by other aspects than their accuracy requirements as it is the case here. Specifically, one could apply this methodology to extract a curriculum from tasks corresponding to reaching distinct regions for a robotic arm, or from tasks distinct in nature and difficulty inside a more complex environment.

The idea of incorporating the final accuracy of the arm movement in the reward scheme of the agent reminds of dense reward functions used in standard versions of Reacher-like environments. In these cases, the shaped reward is proportional to the Euclidean distance between the extremity of the arm and the target position. Yet, we have no guarantee that two target positions close in the Euclidean space can be reached with close policies for complex robotic agents, and thus such shaping could be misleading. On the contrary, in our case, there is no assumption that the image of the control space in the goal space is Euclidean as the generalization from one accuracy to another is learned by the agent through the addition of $\epsilon$ to the input state of \uvfa{}s.

In the future, it would be useful to compare the efficiency of the curriculum obtained from our method with that of a method where a fixed increasing diffulty curriculum is used (see e.g. \citep{clement2013multi,forestier2017intrinsically}). Besides, we intend to compare the explicit form of curriculum based on competence progress used here and the implicit form of curriculum resulting from the \her mechanism. 


\section{Acknowledgments}
This work was supported by the European Commission, within the DREAM project, and has received
funding from the European Unions Horizon 2020 research and innovation program under grant agreement $N^o$ 640891.

\bibliographystyle{agsm}

\end{document}